# Evaluation of General Large Language Models in Contextually Assessing Semantic Concepts Extracted from Adult Critical Care Electronic Health Record Notes

Darren Liu, Cheng Ding, Delgersuren Bold, Monique Bouvier, Jiaying Lu, Benjamin Shickel, Craig S. Jabaley, Wenhui Zhang, Soojin Park, Michael J. Young, Mark S. Wainwright, Gilles Clermont, Parisa Rashidi, Eric S. Rosenthal, Laurie Dimisko, Ran Xiao, Joo Heung Yoon, Carl Yang, Xiao Hu

*Abstract*— **The field of healthcare has increasingly turned its focus towards Large Language Models (LLMs) due to their remarkable performance. However, their performance in actual clinical applications has been underexplored. Traditional evaluations based on question-answering tasks don't fully capture the nuanced contexts. This gap highlights the need for more in-depth and practical assessments of LLMs in real-world healthcare settings.**

**Objective: We sought to evaluate the performance of LLMs in the complex clinical context of adult critical care medicine using systematic and comprehensible analytic methods, including clinician annotation and adjudication.**

**Methods: We investigated the performance of three general LLMs in understanding and processing real-world clinical notes. Concepts from 150 clinical notes were identified by MetaMap and then labeled by 9 clinicians. Each LLM's proficiency was evaluated by identifying the temporality and negation of these concepts using different prompts for an in-depth analysis.**

**Results: GPT-4 showed overall superior performance compared to other LLMs. In contrast, both GPT-3.5 and text-davinci-003 exhibit enhanced performance when the appropriate prompting strategies are employed. The GPT family models have demonstrated considerable efficiency, evidenced by their cost-effectiveness and time-saving capabilities.**

**Conclusion: A comprehensive qualitative performance evaluation framework for LLMs is developed and operationalized. This framework goes beyond singular performance aspects. With expert annotations, this methodology not only validates LLMs' capabilities in processing complex medical data but also establishes a benchmark for future LLM evaluations across specialized domains.**

## I. Introduction

General Large Language Models (LLMs) like GPT-4[1] have demonstrated human-level natural language processing (NLP) capacities and only require text-based prompting to obtain excellent results, lowering the barrier to high-quality text analysis. In healthcare, the potential applications of LLMs are substantial[2-5]. However, it remains unclear if the excellent performance of LLMs in controlled evaluations, like medical examination question responses, can be extrapolated to nuanced clinical scenarios.

In clinical practice, healthcare providers focus on the identification and treatment of disease states by integrating myriad and complex data, such as the timing of a symptom's onset or the presence or absence of specific clinical signs. These details are crucial for accurate diagnosis and treatment planning but are often lacking in benchmark datasets used for testing LLMs. For example, understanding the nuanced difference between a symptom that is persistently present or chronic versus one that is episodic or acute can significantly alter the clinical interpretation and subsequent medical decision-making. In the context of machine learning, NLP with LLMs is a promising approach to capturing such details. However, they may be lacking in training datasets.

Moreover, clinical environments demand a level of adaptability and contextual understanding that goes beyond recalling medical facts. Providers often deal with incomplete information, patient-specific variables, and the need to make quick decisions based on a combination of clinical expertise and patient data. In such scenarios, the ability of an LLM to integrate and interpret this multifaceted information becomes critical.

This calls for a more nuanced and targeted approach to evaluating LLMs, incorporating real-world clinical scenarios and testing the models' abilities to handle the complexities and subtleties of actual medical practice.

To tackle these concerns, our study employs a two-step approach. First, we map free biomedical text to standardized medical concepts aligned with UMLS using MetaMap. This step is crucial for establishing a common ground of understanding between the unstructured clinical text and the structured medical knowledge. Next, we assess each LLM's ability to accurately identify the timing and negation of these medical concepts. This evaluation is carried out using a variety of prompts designed for an in-depth analysis of the models' capabilities. We conducted extensive experiments on a newly curated dataset featuring 2,288 clinical concepts annotated within 150 clinical notes by a team of multidisciplinary clinicians. This dataset not only serves as a testbed for our current study but also stands as a valuable resource for future research in this field. Through this comprehensive approach, our study aims to bridge the gap between LLMs' theoretical performance and their practical utility in the intricate and demanding environment of clinical practice.

## II. METHODS

### A. Data Sources

In this study, we incorporated clinical notes obtained from the Medical Information Mart for Intensive Care III (MIMIC-III) database[6]. Given that nursing notes are recorded at least once a shift, we posited that they are ideal sources to extract clinical concepts that may reflect documentation of acute changes in patients requiring timely recognition. Therefore, we randomly selected 160 clinical notes coded as nursing notes from the MIMIC-III database for analysis.

### B. Concept Extraction

We used MetaMap 2020a[7] to extract clinical concepts from the notes selected for analysis. MetaMap returns the location of a trigger word/phrase in the text associated with a standard clinical concept. Based on a triggering text, we constructed a prompt requesting an evaluation of the corresponding concept as detailed in Section II G. For this study, we only tested concepts from three semantic groups: diseases/syndromes, signs/ symptoms, and mental/behavioral dysfunctions. These semantic types were chosen because such information is often missing from structured EHR data.

### C. Annotation of Extracted Concepts

We enlisted a diverse team of clinicians as annotators, comprised of three nursing researchers and six critical care physicians. We used a subset of 10 notes to first develop our annotation protocol. Each concept was annotated with regard to three questions:
1. Is the concept correctly detected by MetaMap?
2. Is the correctly detected concept being dealt with in the current encounter?
3. Does a correctly detected condition need to be negated (i.e., has the condition already been treated)?

Each extracted concept from the remaining 150 notes comprising the analytic subset was annotated by one nurse and one physician.

Based on the annotation results, two different datasets were generated: 1) dataset A where concepts are concurred by the two experts (concurred group), and 2) dataset B where the two experts disagree (dissenting group). When the two experts did not agree on the classification of a given concept, a nurse researcher (MB) and a physician (CSJ) conducted a final review and adjudicated all concepts from dataset B.

### D. A Web-based Tool for Annotation and Adjudication

Annotation represents a critical phase in numerous machine learning workflows. Therefore, the availability of a user-friendly annotation tool, designed for non-technical experts, becomes imperative. As part of an active effort to develop a large AI-ready dataset for meeting challenges in critical care under the nascent NIH Bridge2AI program, our team has been developing a web-based application Cohort Adjudication and Data Annotation (CADA) targeting many use cases. The application of CADA to support clinical note annotation in this study is an initial demonstration of the intended usage of CADA. A screenshot of CADA is shown in Figure 1. CADA is designed to facilitate the assignment of annotation tasks, enable end-user review of assigned notes and tagged concepts, manage the annotation progress, and facilitate collections of results.

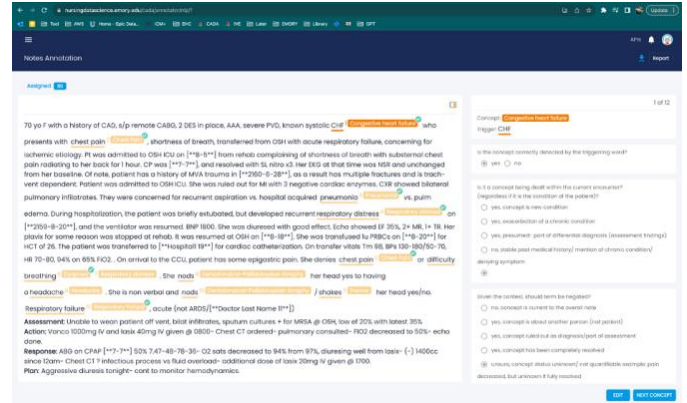

Fig. 1. A screenshot of the annotation and adjudication tool - CADA.

As an annotator in CADA, a user will first see an annotation panel consisting of two main subpanels. The left subpanel displays notes assigned to the user. Trigger words detected by MetaMap are also highlighted in the note with corresponding concepts and three labeling tasks are displayed on the right subpanel. To facilitate obtaining consistent results, we present questions under each labeling task alongside a list of choices. For each posed question, we delineate choices to support annotators in selecting the appropriate responses based on their clinical judgment. We developed a green badge on the top right corner which indicates whether a concept has been fully annotated or not. This allowed annotators to have a clearer view of their progress with their assigned notes.

As an adjudicator, a user will first see a list of notes to be adjudicated, which is sortable by degree of disagreement. All these notes or a subset of them can be selected to be adjudicated using a slightly modified Annotation Panel where the initial annotations are also displayed to inform adjudicators. In this study, the two adjudicators perform adjudication simultaneously note by note.

### E. Selection and Tasking of LLMs

For our study, we selected four general-purpose LLMs to conduct the tagging tasks: text-davinci-003, GPT-3.5, GPT-4, and LLaMA 2. GPT-3.5, GPT-4, and text-davinci-003 are all GPT models from OpenAI and are readily accessible via Microsoft Azure OpenAI API, which meets the privacy and data use requirements for MIMIC-III access. We deployed LLaMA 2, an open-source LLM, on-premises and selected the 7b parameter version recognizing tradeoffs between performance and computational efficiency.

The four selected LLMs were similarly tasked to assess the medical concepts detected by MetaMap by responding to the same three labeling questions posted to expert annotators. We conducted a series of experiments comparing the selected LLMs across different prompts. The overall workflow from data acquisition to the generation of task-specific prompts is outlined in Figure 2.

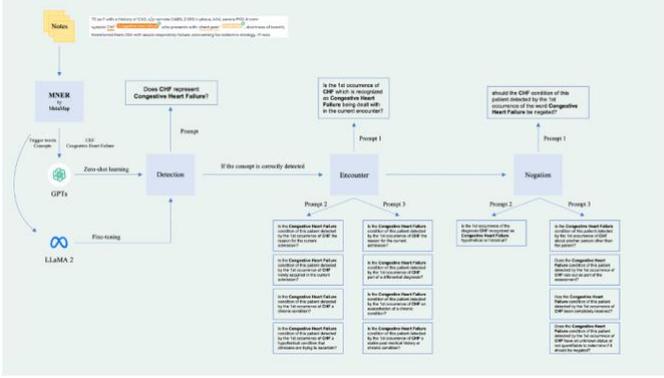

Fig. 2. The flow diagram of the workflow including the prompts we used for all tasks.

*F. GPT Family LLM Prompt Engineering*

The GPT LLM family features a system prompting capability, allowing for the assignment of specific roles with tailored responses. We prompted GPT as follows: "You are a helpful assistant that detects medical concepts from patients' clinical notes. You should only answer 'yes', 'no' or 'unclear'." This approach was selected to promote a consistent and concise response format.

*G. LLM Prompting Approaches for Specific Tasks*

In this study, we focused on experiments involving zero-shot learning. This approach entailed interacting with LLMs without providing any supplementary information, examples, or specific training. We also explored different approaches to prompt for two of the three tasks, as described subsequently.
1) Task 1: Concept Detection

    Evaluating whether a concept was correctly identified is the most straightforward of the three tasks. We employed a single prompt: does (trigger word) represent (concept)?
2) Task 2: Concept Pertinence (or Relevance) to the Encounter

    We initially devised a concise prompt for the task of assessing whether a concept was pertinent to the clinical encounter, akin to what we employed for the detection label (prompt 1). This approach required modification as ambiguity around the word "encounter" was observed to adversely impact performance. We then designed a series of sub-questions (prompt 2) to be presented sequentially as shown in Figure 2.

    Through these more detailed inquiries, we prompted LLMs to provide more complete information about the concept and its relevance to the encounter to further inform the approach to prompting and evaluating alignment with clinician judgment. For example, concepts that are part of a differential diagnosis or that represent an exacerbation of a chronic condition would be adjudicated by clinicians as pertinent to the encounter but not necessarily by an LLM. These additional scenarios were addressed in the final prompting approach (prompt 3) to ensure consistency in the concept evaluation criteria between clinicians and LLMs.
3) Task 3: Concept Negation

    Like the previous two tasks, the first prompt for the negation labeling task was also straightforward consisting of one question (prompt 1). Similarly, we extended the prompts in two ways. First, we continued to pose a single question (prompt 2) but introduced two situations that qualified as "negation". Finally, we incorporated more comprehensive sub-questions and presented a scenario where the concept's status might be uncertain or where it may be challenging to ascertain whether negation is applicable (prompt 3).

*H. Fine-tuning*

We selected LLaMA-2-7B[1] (Vicuna variant) as the LLM model to be fine-tuned since it is an open-source model that can be fully controlled locally. In LLaMA-2-7B, the size of overall trainable parameters is 7 billion. Therefore, full-parameter fine-tuning is not practical given the relatively small size (150) of annotated clinical notes and we selected the low-rank adaptation (LoRA)[8] technique for efficiently fine-tuning the LLM. Using LoRA, we seek to optimize the following objective function:

$$\max_{\Delta\Phi} \sum_{X \in X_{train}} \sum_{t=1}^{|X|} log(Pr_{\Theta+\Delta\Phi}(x_t|\boldsymbol{x}_{1:t-1}))$$

where $\Theta$ denotes the full trainable parameters of the LLM, X_train denotes the whole training samples, and $\boldsymbol{X} = x_{1:t} = \{x_1, x_2, \cdots, x_t\}$ denotes one specific training sample which is essentially a sequence of tokens, and $\Delta\Phi \in \mathbb{R}^{d \times k}$ is a low rank matrix added to each layer of the LLM that will be optimized instead of $\Theta$. In our experiments, we set the rank of $\Delta\Phi$ to be 8, thus leading the size of tunable parameters to be around 4 million (0.06% of overall parameters).

Regarding the format of fine-tuning samples, we follow the concept of instruction-tuning[9] to enable adapting LLM to domain-specific tasks. Unlike the conventional fine-tuning that involves changing the model architecture (typically the output layer), instruction-tuning for LLM only requires instruction-style samples without changing the LLM architecture. The instruction-style samples are essentially a two-round conversation mimicking (1) a human asking an LLM a question of interest; (2) an LLM responding to the question. We make necessary changes to the prompts used for ChatGPT as the instruction tuning data.

*I. Qualitative Assessment of GPT-4 Output*

To further investigate GPT's reasoning process on clinical concepts, we evaluated a random sampling of 300 GPT-generated responses from GPT-4, which was selected owing to favorable performance as reported subsequently. Sixty of the responses were from the detection labeling task, 120 were from the encounter labeling task, and the remaining 120 were from the negation labeling task. The evaluation encompassed six criteria: 1) factuality of the response by GPT; 2) relevance of the GPT-chosen facts to the original questions; 3) completeness of responses; 4) degree of logic (logicality) in the reasoning process by GPT; 5) clarity of the reasoning by GPT; 6) overall human comprehensibility to the response of GPT. We aimed to evaluate GPT-4's performance in collecting data through the first three criteria, in reasoning using the collected data through the next two criteria, and finally in presenting results through the last criterion, offering a comprehensive understanding of its strengths and weaknesses.

The GPT-generated responses were assigned to annotators sequentially. Each annotator was randomly assigned 50 responses in each instance until all 300 responses were delegated. Although any given response was assigned to more than one annotator, we did not expect concordance between annotators for a given response. Each response was therefore considered separately in the analyses.

*J. Statistical Analysis*

The performance of LLMs for various tasks was assessed by the F1 scores, which were calculated by comparing outputs from the LLMs with ground truth annotations by clinicians. To statistically compare the performance across different LLMs, we employed ANalysis Of VAriance (ANOVA) tests. With significant factors and/or interaction effects ($α ≤ 0.05$), a post-hoc Simple Main Effects Analysis with pairwise Student's t-test and Bonferroni correction for multiple comparisons was conducted to further delineate performance differences within each factor. We employed bootstrapping to estimate variances of performance metrics given the cost of repeated LLM experiments at scale. We conducted 50 iterations, each involving a random sampling of 80% of the data with replacement. All statistical analyses were carried out using MATLAB (ver. R2022, MathWorks, Natick, MA).

*K. Code Availability*

The underlying code for this study is available in LLM-evaluation and can be accessed via this link: https://github.com/hulab-emory/LLM-evaluation.

## III. RESULTS

*A. Extracted Concept Annotation Results*

Of the 160 extracted notes, 104 of them were consistent with nursing notes as expected, and the remaining 56 were originally classified as nursing notes but contained long-form narrative content more consistent with documentation by non-nurse clinicians. On average, each note encompassed approximately 44 sentences and 463 words. A total of 2,288 concepts were identified from the 150 notes used for analyses, including 356 unique trigger words, and 276 unique concepts. The distributions of the most frequent concepts and trigger words are shown in Figure 3.

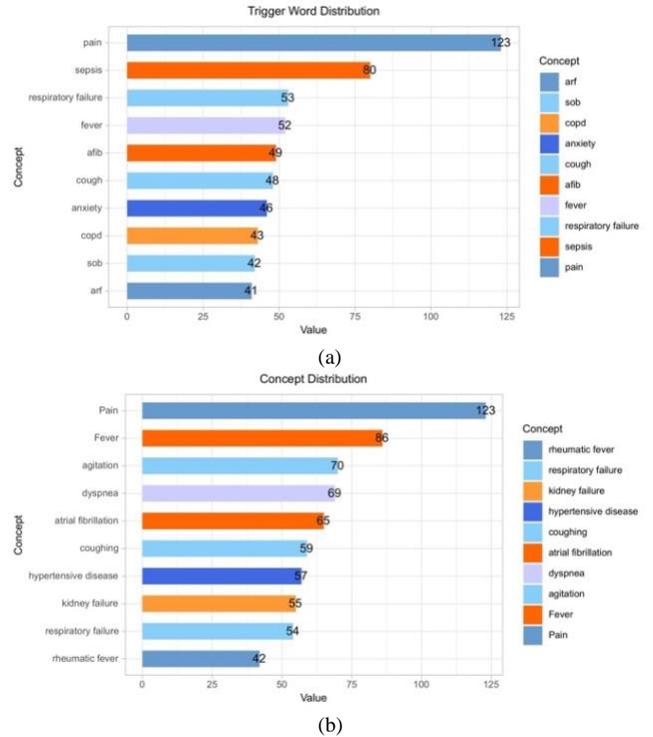

Fig. 3. Distributions of: (a) trigger words; (b) concepts.

Within the 2,288 identified concepts, 2,239 were subjected to dual annotation: one by a nursing researcher and the other by a physician. The annotation team reached an agreement on the detection label for 1,949 concepts with 87.0% inter-rater agreement. For the 1,618 correctly detected concepts, 1,243 were judged by two annotators as pertinent to the encounter with 76.8% inter-rater agreement. Annotators mutually agreed upon the negation task for 1,271 out of 1,618 with 78.6% inter-rater agreement. Overall, 1,441 concepts were agreed upon by both annotators across all labels (concurred group), and the annotators disagreed about one more task for the remaining 798 concepts (dissenting group), all of which were ultimately adjudicated by a nurse researcher and physician.

## B. Comparisons Between LLMs on Different Annotation Datasets

LLM performance was compared against clinician adjudicators in up to three tasks for each of the extracted concepts as outlined in Section II C. In the case of the encounter and negation labeling tasks, a concept was only included in the final performance metrics when LLMs correctly classified it as being appropriately detected by MetaMap, which means LLM performance across the encounter and negation labeling tasks was only evaluated for concepts where clinicians and LLMs agreed that a concept was appropriately identified by MetaMap.

From the results shown in Table 1, it is evident that LLMs generally exhibit superior performance on the concured dataset. Particularly in the encounter labeling task, all tested LLMs almost consistently achieved F1 scores exceeding 0.9 on the concured dataset. However, on the dissenting dataset, none of the models attained an F1 score over 0.9, regardless of the prompt used. Notably, an exception was observed with GPT-3.5, which demonstrated better F1 scores on the dissenting dataset for the negation labeling task, specifically when engaged with prompt 3 (i.e., the prompt with 4 sub-questions). These findings suggest that the concured dataset, while being more straightforward for human experts to analyze, also presents a less challenging environment for LLMs.

When comparing different LLMs, we observed that for each task, both GPT-4 and text-davinci-003 surpassed an F1 score of 0.9 when evaluated with their best-performing prompt on the concured dataset. GPT-3.5 did not achieve a comparably high F1 score in the negation labeling task. It has been observed that GPT-4, as the latest and most advanced model in this series, requires minimal prompt engineering to achieve high performance. When comparing performance between different prompting approaches to a given task, GPT-4 demonstrated consistently high performance. While GPT-3.5 and test-davinci-003 are both GPT-3 fine-tuned models, it is interesting to see the performance difference between GPT-3.5 and text-davinci-003, particularly with GPT-3.5 exhibiting significantly lower performance in many cases. This observation suggests that fine-tuning the model with a focus on text completion, as opposed to conversational capabilities, enhances its effectiveness in our specific tasks.

For each task, both a three-way ANOVA test (or a two-way ANOVA test for the detection labeling task since only one prompt was used) and a post-hoc Simple Main Effects Analysis were conducted. The results shown in Tables S1 to S3 demonstrate that the two-way/three-way ANOVA tests identified significant differences in F1 scores across 3 tasks, 3 different LLMs, and 2 datasets, with notably small p-values ($p \ll 0.05$). In the post-hoc analysis, employing pairwise Student's t-tests with Bonferroni correction, it was found that all LLMs exhibited significantly different performances on the two datasets. The only exception was the lack of a significant performance difference between GPT-3.5 and GPT-4 for prompt 3 on the concured dataset in the encounter labeling task.

As indicated in Table 1 and Figure 4, the standard deviation is considerably higher for the dissenting dataset compared to the concured dataset. This further corroborates the observation that LLMs exhibit a similarly increased variance in their performance, akin to human experts when dealing with the more challenging dissenting dataset.

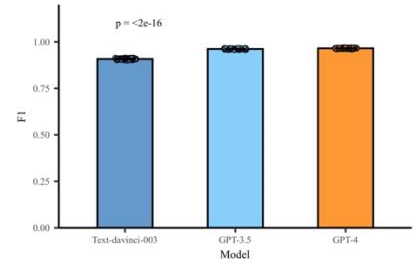
(a)

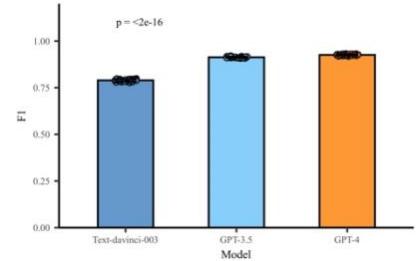
(b)

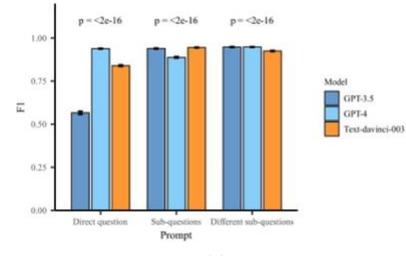
(c)

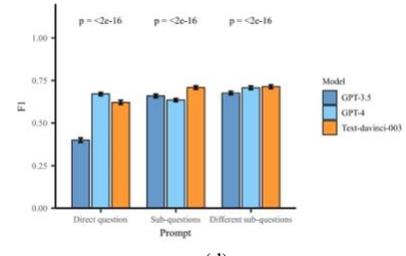
(d)

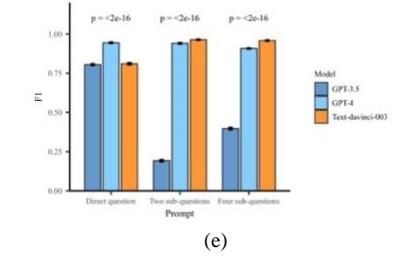
(e)

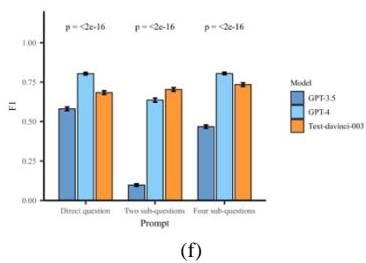

(f)

Fig. 4. F1 scores of different LLMs on different labeling tasks and datasets: (a) detection task on dataset A; (b) detection labeling task on dataset B; (c) encounter labeling task on dataset A; (d) encounter labeling task on dataset B; (e) negation labeling task on dataset A; (f) negation labeling task on dataset B.

*C. Comparison Fine-Tuned LLaMA 2 with Matching Test Data Using the First Prompt of All Tasks*

Table 2 reveals that GPT-4 maintains consistently high performance across all tasks, excelling particularly in the detection and negation labeling tasks with F1 scores of 0.948 and 0.887, respectively. In the encounter labeling task, while zero-shot LLaMA 2 outperforms others, GPT-4 still achieves a comparable F1 score. Notably, although LLaMA 2 initially shows weaker performance in the detection and negation labeling tasks with zero-shot learning, fine-tuning significantly enhances its performance, elevating it to a level comparable with GPT-4, thereby making it the second-best model in these tasks. However, when fine-tuned, LLaMA 2's F1 score slightly diminishes to 0.853.

The analysis of standard deviations reveals that text-davinci-003, when using zero-shot learning, shows notably higher variance, with a significant standard deviation of 0.209 for the encounter labeling task. In contrast, GPT-4 demonstrates the most consistent results, with all its standard deviations remaining below 0.007. Surprisingly, LLaMA 2 with zero-shot learning reaches a sensitivity of 1 and a standard deviation of 0 in the encounter labeling task. This uniformity in sensitivity, indicating that all its responses were positive, is due to the absence of negative responses from LLaMA 2 in this specific task.

A two-way ANOVA test and a post-hoc Simple Main Effects Analysis were also conducted on each task. As indicated in Table S4, the two-way ANOVA test, with LLMs and tasks as the factors and model performance as the response variable, shows significant differences in performance across 3 LLMs (F-test = 21,077.8; $p<0.05$) and the three tasks (F-test = 58,338.46; $p<0.05$), and there is a significant interaction effect between the two factors (F-test = 14,713.05; $p<0.05$). The post-hoc Simple Main Effects Analysis shows that all LLMs achieve significantly different performances except that no significant difference is detected between text-davinci-003 and LLaMA 2 zero-shot learning for the detection labeling task ($p>0.005$), and GPT-4 and LLaMA 2 zero-shot learning for the encounter labeling task ($p>0.005$).

*D. Qualitative Assessment of GPT-4's Output Across Six Criteria*

Within the 378 annotation values we received, 78 were related to the detection labeling task, 153 were related to the encounter labeling task, and 147 were to the negation labeling task. Responses from GPT-4 to 65.9% of these tasks were rated as perfect across all six criteria, and none of the responses from GPT-4 were ranked lowest across all criteria. In Table 3, we break down the percentage of GPT-4 responses falling in each rating of the six criteria and different labeling tasks, which can be summarized as follows.

1) **Factuality:** factuality evaluates if GPT-4's responses rely on factual evidence or if they generate content that is not present in the notes to justify their conclusions. Based on the annotations across all three tasks, 81.7% of GPT-4's responses were wholly grounded in factual evidence (positive). Conversely, 9.0% of the responses were indicated to be only partially reliant on factual evidence (neutral), while the remaining 9.3% of the responses lacked any factual basis (negative). The criterion of factuality exhibited the highest proportion of negative ratings compared to the other evaluated criteria.

2) **Relevance:** relevance examines whether GPT-4 chooses pertinent evidence from the notes to bolster its conclusions and assesses if the overall responses from GPT-4 align with the questions posed. Based on the annotations across all three tasks, 84.9% of GPT-4's responses were completely relevant. In contrast, 12.2% of the responses were deemed to be of partial relevance, and only 2.9% of the responses were viewed as entirely irrelevant.

3) **Completeness:** completeness evaluates the extent to which GPT-4's responses incorporate all the evidence from the notes that can support its conclusions. In this category, 71.7% of GPT-4's responses were fully complete. Meanwhile, 24.9% of the responses were only somewhat complete, and only 3.4% of the responses were highly incomplete. The criterion of completeness exhibited the highest proportion of neutral ratings compared to the other evaluated criteria.

4) **Logicality:** logicality assesses the coherence of GPT-4's reasoning. This criterion evaluates whether the evidence selected by GPT-4 logically substantiates its conclusions and whether the conclusions drawn are correct. Based on the annotations, 79.9% of the responses from GPT-4 were completely logical. Conversely, 12.4% of the responses were partially logical, containing some logical discrepancies, while 7.7% of the responses were deemed entirely illogical.

Table 1. Performance metrics across all tasks, models, and prompts.

| Task | Model | Prompt | Dataset | Accuracy | F1 | PPV | Sensitivity |
|---|---|---|---|---|---|---|---|
| **Detection** | Text-davinci-003 | | Concurred | 0.867±0.004 | 0.908±0.003 | 0.966±0.003 | 0.857±0.005 |
| | | | Dissenting | 0.701±0.007 | 0.789±0.006 | 0.911±0.006 | 0.696±0.008 |
| | GPT-3.5 | | Concurred | 0.941±0.003 | 0.962±0.002 | 0.959±0.003 | 0.965±0.003 |
| | | | Dissenting | 0.857±0.006 | 0.913±0.004 | 0.892±0.005 | 0.936±0.006 |
| | GPT-4 | | Concurred | 0.947±0.003 | 0.966±0.002 | 0.967±0.003 | 0.964±0.003 |
| | | | Dissenting | 0.879±0.006 | 0.926±0.004 | 0.915±0.006 | 0.937±0.005 |
| **Encounter** | Text-davinci-003 | 1 | Concurred | 0.726±0.007 | 0.839±0.004 | 0.909±0.005 | 0.779±0.007 |
| | | | Dissenting | 0.501±0.012 | 0.621±0.012 | 0.522±0.014 | 0.768±0.014 |
| | | 2 | Concurred | 0.899±0.005 | 0.945±0.003 | 0.946±0.004 | 0.944±0.004 |
| | | | Dissenting | 0.604±0.011 | 0.709±0.010 | 0.584±0.013 | 0.903±0.008 |
| | | 3 | Concurred | 0.865±0.005 | 0.925±0.003 | 0.936±0.004 | 0.914±0.004 |
| | | | Dissenting | 0.611±0.011 | 0.714±0.010 | 0.594±0.013 | 0.894±0.010 |
| | GPT-3.5 | 1 | Concurred | 0.441±0.009 | 0.566±0.009 | 0.968±0.004 | 0.400±0.009 |
| | | | Dissenting | 0.598±0.009 | 0.400±0.013 | 0.795±0.017 | 0.267±0.010 |
| | | 2 | Concurred | 0.891±0.005 | 0.939±0.003 | 0.952±0.003 | 0.926±0.004 |
| | | | Dissenting | 0.567±0.011 | 0.659±0.010 | 0.545±0.012 | 0.835±0.011 |
| | | 3 | Concurred | 0.905±0.005 | 0.948±0.003 | 0.945±0.004 | 0.950±0.003 |
| | | | Dissenting | 0.589±0.011 | 0.677±0.010 | 0.547±0.012 | 0.887±0.010 |
| | GPT-4 | 1 | Concurred | 0.886±0.004 | 0.938±0.003 | 0.915±0.004 | 0.964±0.003 |
| | | | Dissenting | 0.565±0.010 | 0.671±0.010 | 0.538±0.011 | 0.891±0.009 |
| | | 2 | Concurred | 0.814±0.007 | 0.888±0.004 | 0.970±0.003 | 0.819±0.007 |
| | | | Dissenting | 0.617±0.008 | 0.635±0.009 | 0.607±0.012 | 0.666±0.012 |
| | | 3 | Concurred | 0.905±0.004 | 0.948±0.002 | 0.935±0.003 | 0.961±0.003 |
| | | | Dissenting | 0.617±0.011 | 0.707±0.010 | 0.571±0.012 | 0.929±0.008 |
| **Negation** | Text-davinci-003 | 1 | Concurred | 0.708±0.008 | 0.810±0.007 | 0.949±0.005 | 0.708±0.008 |
| | | | Dissenting | 0.687±0.011 | 0.683±0.011 | 0.683±0.014 | 0.687±0.011 |
| | | 2 | Concurred | 0.975±0.002 | 0.963±0.004 | 0.951±0.004 | 0.975±0.002 |
| | | | Dissenting | 0.795±0.008 | 0.704±0.011 | 0.631±0.013 | 0.795±0.008 |
| | | 3 | Concurred | 0.972±0.003 | 0.958±0.004 | 0.944±0.006 | 0.972±0.003 |
| | | | Dissenting | 0.817±0.008 | 0.734±0.011 | 0.667±0.013 | 0.817±0.008 |
| | GPT-3.5 | 1 | Concurred | 0.700±0.007 | 0.805±0.005 | 0.946±0.005 | 0.700±0.007 |
| | | | Dissenting | 0.558±0.010 | 0.581±0.011 | 0.796±0.008 | 0.558±0.010 |
| | | 2 | Concurred | 0.113±0.004 | 0.192±0.007 | 0.928±0.009 | 0.113±0.004 |
| | | | Dissenting | 0.065±0.005 | 0.098±0.008 | 0.453±0.031 | 0.065±0.005 |
| | | 3 | Concurred | 0.257±0.006 | 0.397±0.007 | 0.932±0.007 | 0.257±0.006 |
| | | | Dissenting | 0.355±0.009 | 0.468±0.010 | 0.695±0.012 | 0.355±0.009 |
| | GPT-4 | 1 | Concurred | 0.955±0.003 | 0.944±0.004 | 0.935±0.005 | 0.955±0.003 |
| | | | Dissenting | 0.840±0.007 | 0.804±0.008 | 0.800±0.008 | 0.840±0.007 |
| | | 2 | Concurred | 0.943±0.004 | 0.941±0.004 | 0.939±0.005 | 0.943±0.004 |
| | | | Dissenting | 0.695±0.010 | 0.636±0.013 | 0.604±0.014 | 0.695±0.010 |
| | | 3 | Concurred | 0.870±0.005 | 0.908±0.004 | 0.950±0.004 | 0.870±0.005 |
| | | | Dissenting | 0.798±0.008 | 0.805±0.008 | 0.820±0.008 | 0.798±0.008 |

Table 2. Performance metrics comparing zero-shot learning and fine-tuning.

| Task | Model | Accuracy | F1 | PPV | Sensitivity |
|---|---|---|---|---|---|
| **Detection** | Text-davinci-003 (zero-shot) | 0.819±0.044 | 0.874±0.034 | 0.941±0.010 | 0.816±0.072 |
| | GPT-3.5 (zero-shot) | 0.907±0.004 | 0.941±0.002 | 0.923±0.003 | 0.960±0.003 |
| | GPT-4 (zero-shot) | 0.919±0.003 | 0.948±0.002 | 0.940±0.003 | 0.957±0.002 |
| | LLaMA 2 (zero-shot) | 0.780±0.005 | 0.874±0.003 | 0.782±0.005 | 0.989±0.002 |
| | LLaMA 2 (fine-tuned) | 0.889±0.003 | 0.929±0.002 | 0.918±0.004 | 0.940±0.003 |
| **Encounter** | Text-davinci-003 (zero-shot) | 0.674±0.086 | 0.795±0.130 | 0.798±0.073 | 0.792±0.209 |
| | GPT-3.5 (zero-shot) | 0.504±0.008 | 0.535±0.008 | 0.944±0.006 | 0.374±0.008 |
| | GPT-4 (zero-shot) | 0.767±0.007 | 0.861±0.005 | 0.791±0.007 | 0.946±0.004 |
| | LLaMA 2 (zero-shot) | 0.761±0.007 | 0.864±0.005 | 0.761±0.007 | 1 |
| | LLaMA 2 (fine-tuned) | 0.751±0.007 | 0.853±0.004 | 0.777±0.007 | 0.946±0.004 |
| **Negation** | Text-davinci-003 (zero-shot) | 0.708±0.027 | 0.769±0.026 | 0.848±0.028 | 0.708±0.027 |
| | GPT-3.5 (zero-shot) | 0.655±0.007 | 0.718±0.007 | 0.897±0.017 | 0.655±0.007 |
| | GPT-4 (zero-shot) | 0.911±0.004 | 0.887±0.006 | 0.875±0.006 | 0.911±0.004 |
| | LLaMA 2 (zero-shot) | 0.565±0.007 | 0.645±0.006 | 0.776±0.007 | 0.565±0.007 |
| | LLaMA 2 (fine-tuned) | 0.839±0.005 | 0.792±0.007 | 0.763±0.008 | 0.839±0.005 |

5) **Clarity:** clarity indicates whether GPT-4's statements are clear enough for clinicians to understand. Based on the annotations across all tasks, 85.2%, 10.8%, and 4.0% of the responses were deemed completely clear, moderately clear, or utterly unclear, respectively.
6) **Overall Comprehensibility:** while the previous 5 criteria delve into specific attributes, this criterion evaluates whether GPT-4's responses are comprehensible overall. 85.4% of the responses were completely comprehensible, 13.0% of the responses were deemed to be somewhat comprehensible, and only 1.6% of the responses from GPT-4 were incomprehensible.

Table 3. Evaluations of GPT-4 on six criteria

| Task | Value | Count % |
|---|---|---|
| Detection | Completely factual | 85.9 |
| | Somewhat factual | 10.3 |
| | Not factual | 3.8 |
| Encounter | Completely factual | 70.0 |
| | Somewhat factual | 18.3 |
| | Not factual | 11.8 |
| Negation | Completely factual | 91.8 |
| | Somewhat factual | 5.4 |
| | Not factual | 2.7 |

(a)

| Task | Value | Count % |
|---|---|---|
| Detection | Completely relevant | 85.9 |
| | Somewhat relevant | 9.0 |
| | Not relevant | 5.1 |
| Encounter | Completely relevant | 77.8 |
| | Somewhat relevant | 18.3 |
| | Not relevant | 3.9 |
| Negation | Completely relevant | 91.9 |
| | Somewhat relevant | 7.5 |
| | Not relevant | 0.7 |

(b)

| Task | Value | Count % |
|---|---|---|
| Detection | Completely complete | 75.6 |
| | Somewhat complete | 23.1 |
| | Not complete | 1.3 |
| Encounter | Completely complete | 60.8 |
| | Somewhat complete | 33.3 |
| | Not complete | 5.9 |
| Negation | Completely complete | 81.0 |
| | Somewhat complete | 17.0 |
| | Not complete | 2.0 |

(c)

| Task | Value | Count % |
|---|---|---|
| Detection | Completely logical | 83.3 |
| | Somewhat logical | 11.5 |
| | Not logical | 5.1 |
| Encounter | Completely logical | 69.3 |
| | Somewhat logical | 17.0 |
| | Not logical | 13.7 |
| Negation | Completely logical | 89.1 |
| | Somewhat logical | 8.2 |
| | Not logical | 2.7 |

(d)

| Task | Value | Count % |
|---|---|---|
| Detection | Completely clear | 84.6 |
| | Somewhat clear | 12.8 |
| | Not clear | 2.6 |
| Encounter | Completely clear | 79.7 |
| | Somewhat clear | 13.7 |
| | Not clear | 6.5 |
| Negation | Completely clear | 91.1 |
| | Somewhat clear | 6.8 |
| | Not clear | 2.0 |

(e)

| Task | Value | Count % |
|---|---|---|
| Detection | Completely comprehensible | 87.2 |
| | Somewhat comprehensible | 12.8 |
| | Not comprehensible | 0.0 |
| Encounter | Completely comprehensible | 79.1 |
| | Somewhat comprehensible | 17.6 |
| | Not comprehensible | 3.3 |
| Negation | Completely comprehensible | 91.2 |
| | Somewhat comprehensible | 8.2 |
| | Not comprehensible | 0.6 |

(f)

## E. Annotation Cost and Time Comparative Analysis

Annotation platform usage data showed that the complete annotation process for 150 notes required approximately 50.8 hours, with each of our 9 annotators contributing an average of 5.6 hours. Considering that each concept was reviewed by two annotators, it would require around 25.4 hours for one human expert to complete all the annotations (40.0s per concept). Given competing demands, the entire annotation process, including the development of the protocol, annotating 150 notes, and adjudicating all disagreements, spanned approximately 3 months.

In contrast, LLMs required 1.4 hours (2.2s per concept) to complete all tasks, approximately 18 times more efficient than human experts. Running all prompts on all 3 GPT models through the Azure OpenAI API incurred a total cost of $5470.49, of which $4580.52 was expended on GPT-4 alone. This averaged $455.87 per complete annotation process, whereas human experts incur a significantly higher cost of approximately $2047.40 (calculated based on HR-reported national average salary of physicians and nurses in the US). Since we were running and fine-tuning LLaMA 2 locally, it is not factored into the cost calculations provided here. Regarding attentiveness and repetitiveness, conceptually human performance tends to decline with repetition whereas model performance and attentiveness improve; however, we did not directly study these effects.

## IV. Discussion

Clinical concept extraction and understanding from notes have been a canonical NLP task[10-13]. Extracted concepts coupled with contextual understanding can be used to support many downstream tasks of clinical relevance, such as predicting various patient outcomes[14-16]. The value of clinical notes lies in the fact that they capture nuances of patient care such as clinicians' observations of subtle signs and symptoms at the bedside that are not available in structured EHR data. While there have been an increasing number of publications investigating the potential of general LLMs in clinical medicine[17-20], testing their abilities in extracting clinical concepts and providing contextual information on extracted concepts has not been adequately addressed. The present study fills this gap with three contributions. First, concepts extracted from clinical notes were annotated by a team of interdisciplinary clinicians to provide ground for evaluating LLM performance under realistic conditions. Second, clinicians comprehensively evaluated responses from the highest-performing model, GPT-4, in regard to three essential steps of clinical concept understanding: collection of evidence, reasoning over the collected evidence, and presentation of the finding. These qualitative evaluations complement our objective assessment of LLMs. Finally, by leveraging well-developed clinical name entity recognition tools such as MetaMap, our study was able to focus more on testing LLM's ability to understand clinical concepts by testing several prompting strategies, demonstrating the superiority of decomposing questions into clinically informed sub-questions to better structure LLM output.

### A. Related Work

Agrawal's study[21] investigated using prompts for GPT-3.5 to extract information from clinical notes and demonstrated the superior performance of zero-shot general LLM over fine-tuned small models. However, it only tested GPT-3.5 without using complete real-world clinical notes. Subsequent studies further explored the use of general LLMs to analyze clinical data in various specialties. Hu's study[22] tested ChatGPT to extract 11 tumor characteristics from CT reports of lung cancer patients and found that ChatGPT achieved high accuracies (0.88-0.99) with a zero-shot approach. Another study[23] also focused on assessing oncologic progression by analyzing CT reports from lung cancer patients and found that GPT-4 outperformed ChatGPT based on a test dataset of 424 reports with regard to both the accuracies and the oncologic reasoning process. ChatGPT was also tested for a use case in understanding laboratory medicine test results based on simulated laboratory test results[24]. It was found in the study that ChatGPT was able to recognize all tests and if they deviated from reference values. However, ChatGPT could only provide superficial interpretations and cannot prescribe meaningful follow-up procedures implying the limitation of the approach to provide higher level decision support in laboratory medicine. Delsoz's study[25] reported using ChatGPT to analyze case reports to diagnose different forms of glaucoma at a level matching that of senior ophthalmology residents and Boyle's study[26] tested LLaMA 2, GPT-3.5, and GPT-4 to generate ICD codes based on publicly available CodiEsp dataset[27]. The design of prompts is inherently a trial-and-error approach. Therefore, the study[10] systematically compared 6 different styles of prompts on various information extraction tasks from clinical notes on the CASI dataset[28].

### B. The need for real-world evidence of LLM's ability in clinical concept extraction and understanding

In comparison with studies reviewed in Section IV A, our work used a completely new set of clinical notes from the MIMIC database and engaged a multidisciplinary team of clinicians to annotate these notes from scratch. Releasing annotations openly to the research community, our effort provides much-needed resources to further develop and test LLM's role in performing clinically relevant tasks. Furthermore, our pipeline to process MIMIC data can be readily replicated to process more private data at individual institutions.

### C. How Well Did GPT-4 Understand Clinical Concepts?

Our project was largely motivated by the early success of LLMs in handling questions in standardized medical[29] and nursing licensure exams[30]. LLMs have demonstrated superb abilities in reasoning clinical questions without training with comprehensive medical literature[31], a phenomenon that has been reported in other fields[32]. To gain a deeper understanding of the process that LLMs use to perform this high-level task, we framed our qualitative evaluation approach by treating the LLM under evaluation as a virtual assistant tasked with collecting evidence, reasoning with the collected evidence, and eventually describing the findings. GPT-4 was chosen as the LLM to be evaluated in this way owing to its superior objective performance. While previous studies laid a foundational understanding of LLM applications in healthcare, their scope is limited to singular aspects, such as relevance[33] or correctness[29], or focused on choosing the best answer from multiple-choice questions[34,35]. Our approach, in contrast, encompasses a broader spectrum of qualitative performance. This methodology not only highlights the superior capability of GPT-4 in processing complex medical data but also sets a precedent for future research in evaluating LLMs across various specialized domains.

*D. Focus on LLM's Ability to Understand Clinical Concepts*

In comparison with relevant studies reviewed in Section IV A, our approach of using MetaMap to recognize clinical concepts in the target semantic groups first has the advantage of enhancing LLM's ability to understand these concepts with prompts that are already enriched with the supplied concept. This approach necessitates first presenting the "detection" question as MetaMap can make mistakes in recognizing clinical concepts. Our results show that all tested LLMs were able to answer this detection question with high accuracy using a simple one-question prompt, demonstrating the effectiveness of customizing prompts with more contextual information. We also demonstrated that clinically informed prompts with a breakdown of possible clinical scenarios can be better leveraged by advanced LLMs such as GPT-4 to provide accurate, easily interpretable responses pertaining to clinical concepts. We acknowledge that an intriguing consideration would be the possibility of replacing MetaMap entirely with LLMs for named entity extraction tasks. While such an approach may appear to be more "self-contained", it will be challenging to develop more global prompts and strategies to handle the increased chance of hallucinations and ensure consistent, standardized outputs.

*E. Limitations and Future Directions*

The use of single-center data from the MIMIC-III database may limit the generalizability of results to data from other centers. Potential barriers to testing LLMs external to an institution include ensuring no inadvertent leakage of sensitive information in clinical notes as well as thorough security review of chosen systems following institutional policy. At the moment, MIMIC is the only publicly available open dataset that contains fully de-identified clinical notes in their entirety, therefore justifying their use in studies like ours as an initial effort to test state-of-the-art LLMs while efforts are underway to remove the barriers to future studies of notes from a wider array of EHRs. It can be also argued that clinical NLP, as a field, can benefit from more de-identified notes that are openly shared. Interestingly, LLMs may be used as a powerful tool to de-identify clinical notes[36].

Disagreement in clinicians' assessment of clinical concepts implies that our ground truth may carry inherent subjectivity. To alleviate this limitation, we separately reported results that are based on the concured subset of clinical concepts and those that show disagreement in clinicians' annotations. In this way, conclusions regarding the performance of evaluated LLMs can be more convincingly supported by results from the concured subset. However, we cannot exactly pinpoint the reason for a worse performance on the dissenting subset. Some plausible reasons may include the inherent variation due to the subjectivity in the adjudication or more intrinsically difficult concepts in this subset. One solution in the future is to provide a more complete EHR for a given patient to help clinicians annotate clinical concepts informed by additional data.

Additionally, the study did not investigate potential enhancements from instruction-specific fine-tuning of GPT-4. Focusing on zero-shot performance is a reasonable choice to establish a lower-bound performance as a foundation step for future studies. To further improve the performance of LLMs on datasets from a specific healthcare enterprise, several institutions have trained LLMs using their own institutional EHR data[37-39]. These models followed the same recipes of model architecture as used in developing general LLMs but are much smaller than general-purpose LLMs. Because these models lack support for prompt-based interactions with the model, it remains interesting to establish concrete benefits from training organization-specific LLM.

Lastly, the rapidly evolving nature of LLM technology and clinical practices could limit the long-term applicability of the study's findings. Addressing these limitations in future research is essential to enhance the practical application of LLMs in healthcare.

V. CONCLUSION

The present work establishes the superiority of GPT-4, over earlier versions of GPT models and a locally fine-tuned LLaMA-7B model, in extracting and understanding 2,288 clinical concepts in 150 clinical notes in the MIMIC III database that are from three semantic groups: diseases/syndromes, signs/symptoms, and mental/behavioral dysfunctions. Based on the concured dataset, the highest F1 values reached by GPT-4, in a zero-shot fashion, are 0.966, 0.948, and 0.944 for detection, encounter, and negation labeling tasks, respectively. These encouraging results motivate further testing of GPT-4's generalizability in processing data from different EHRs and improving its performance through in-context learning and other more advanced techniques such as retrieval augmented generation.

**Ethics statement.** The study was deemed as not human subjects research by the Emory IRB. We used the Azure OpenAI service in order to comply with the terms of the PhysioNet Credentialed Health Data Use Agreement 1.5.0. Specifically, we explicitly required no external human review of the data and only processed data using a deployment within our control in our Azure cloud environment. All investigators of this project completed HIPAA training at their own institutions and signed the aforementioned data use agreement.

AUTHOR CONTRIBUTIONS

XH and CD conceived the project. DL engineered pipelines for generating responses from GPT models and conducted analyses of the outcomes. CD applied MetaMap to identify and extract trigger words and concepts. DB developed the annotation and adjudication tool CADA and took charge of assigning and assisting annotations and adjudications. JL fine-tuned LLaMA 2 and was responsible for eliciting responses from both the fine-tuned and zero-shot iterations of LLaMA 2. RX executed comprehensive statistical analyses. MB, CJ, WZ, SP, MY, MW, GC, ER, and LD performed annotations. MB and CJ created the annotation protocol with the annotation team and carried out adjudications.

JY, XH, and DL collaboratively authored the abstract. CD crafted the Introduction Section. DL took the lead in writing both the Method and Results Sections. Within the two sections, JL contributed to the fine-tuning subsection, and RX added detailed statistical analysis. XH, JY, CD, and DL finalized the Discussion and Conclusion Sections. MB, BS, CJ, PR, ER, and JY made significant contributions by substantially enhancing and refining the literature.

DL, CD, DB, MB, and JL contributed equally to this project.

## DATA AVAILABILITY

Clinical notes analyzed in the study are available from publicly available MIMIC III database. The unique identifiers of the 160 notes annotated in this study and the annotations results will be made publicly available on PhysioNet and requests for this dataset can be also made to the corresponding author.

## COMPETING INTERESTS

All authors declare no financial or non-financial competing interests.

## ACKNOWLEDGMENT

Research reported in this manuscript was partially supported by the Office of the Director, National Institutes of Health under OT award number OT2OD032701. The content is solely the responsibility of the authors and does not necessarily represent the official views of the National Institutes of Health. SP was additionally supported by NIH R01 NS131606 and NS129760. JY was supported by NIH K23 GM138984.

SUPPLEMENTARY INFORMATION

*Analysis of False Responses by GPT-4*

Case 1 – False Negative
Note:

[Clinical note image: HPI: Mr [**Known lastname 8206**] is a 57 yo male with etoh cirrhosis (Cirrhosis ✓, Liver Cirrhosis ✓), now on transplant list, initially presented with variceal bleed, transferred from [**Hospital **] Medical Center for TIPS evaluation. Hospital course complicated by PNA requiring intubation, hepatic encephalopathy, anemia and most recently fevers, altered mental status. Evaluation of fevers revealed positive UA, CXR c/w pneumonia, and SBP all diagnosed. Fever developed [**9-17**] AM with associated AMS, tachycardia. Pt initially treated wiht CTX but broadened to Zosyn and Vancomycin once PNA, UTI (Urinary tract infection ✓) and SBP all diagnosed. Pt developed hypotension to SBP 70s-80s, from baseline 100s and was given 500cc bolus and ordered 100 albumin. Pt transferred to MICU for close monitoring. Sepsis without organ dysfunction
Assessment: Pt with WBC 2.8, HR in 105-110, temperature 100.1 rectally at the time transfer, BP 70-80 systolically.
Action: Pt initially bolused with fluid without effect, started on Levo and titrated to keep MAP above 65. Given Additional 2L fluid bolus, Cultures from Ascitic fluid as well as BC came back positive with Gram negative rodes,Given Zosyn and Vancomycin as ordered. [**Month/Year (2) 865**] placedon his left radial artery.
Response: Levo currently ay 0.1 mics/hr,MAP above 65,Lactate 7.5,Temp down to 97.9 axillary.
Plan: Monitor vital signs closely, Titrate Levo to keep Map above 65, More fluids?, Follow up with culture and Cont abx. Altered mental status (Abnormal mental state ✓) (not Delirium ✓ Delirium ✓)
Assessment: Likely due to Hepatic encephalopathy, alert disoriented, not following commands, MAE,PERLA 4mm and reactive.
Action: Getting Lactulose Q4h, Cont Rifaximin,Folow up with lever. Tolerating PO meds.
Response: Flexiseal draining golden yellow loose stool. Pt looks comfortable.
Plan: Closely monitor mental status,Cont Lactulose,Follow up with Liver.]

**Question:** Does the word 'Delirium' represent 'Delirium'?
**Answer:** No, the word 'Delirium' does not represent 'Delirium' in this context. The note specifically states "Altered mental status (not Delirium)", indicating that the patient's altered mental status is not due to Delirium.
**Analysis:** The note mentions "Altered mental status (not Delirium)". The phrase "not Delirium" within parentheses suggests that while the patient exhibits an altered mental state, it is specifically not due to Delirium. However, GPT-4 mistakenly considers the 'Not' as the negation for representing Delirium.

Case 2 – False Positive
Note:

[Clinical note image: Pt 82 yo male who is transferred from OSH after presenting to PCP with several days of chest discomfort, congestion, fever, nausea (Nausea) and decreased appetite. Ruled in for NSTEMI, intubated for pul edema at [**Hospital 1991**] transferred to [**Hospital1 1**] for further care. Extubated [**6-21**] then later that day required mask ventilation for increased resp distress, pul edema, given lasix. Has cont to have fever (Fever), CXR with PNA on R, antibiotics broadened. S/P cardiac cath [**6-22**] w/ clean coronaries, PCWP 15, CO 4.8, CI 2.25. Acute coronary syndrome (ACS, unstable angina, coronary ischemia)
Assessment: Denied CP. Stable hemodynamics w/ HR 60s SR, BP 120s/70s. K 3.0 this AM.
Action: *Medical management w/ IV heparin (currently off), po lopressor, ASA, plavix. * K repleted w/ 40mEq KCL PO and 40mEq KCL IV in 500 NS. *To cath lab @ 12:00noon. IV NaHCO3 bolus given 1 hr pre-cath as ordered. *Returned from cath lab @ 15:30 s/p R&L heart cath w/ clean coronaries, PCWP 15, CO 4.79/ CI 2.25. R groin arterial/venous sheaths D/Cd in holding area w/ hemostasis at 15:20. *Heart healthy diet teaching initiated.
Response: Stable hemodynamics post cath. R groin site CDI, no ooze/ hematoma. Easily palp DP/PT bilat. Pts wife asking appropriate questions re: diet teaching.
Plan: Continue to monitor hemodynamics. IV heparin->off during cath. To be restarted at midnight with no bolus. Continue to follow R groin/ pedal pulses. Strict bedrest until 21:20. Needs Post MI teaching reinforcement. Pneumonia, bacterial, community acquired (CAP)
Assessment: LS clear w/ scattered rhonchi/ crackles on R side. SPO2 >95% on face tent @ 50%/ 15L O2. Pt expectorating moderate amts of thick, dark bloody secretions. Tmax this shift 101.9 po. WBC only 6.7.
Action: IV anbx as ordered. Encouraged cough and deep breathing. Given IS and instructed on proper use. Assisted pt w/ oral care q 4 hours. Tylenol 650mg PRN fever.
Response: Pt able to expectorate sputum with current regimen. Temp down w/ Tylenol.
Plan: Continue to monitor fever curve. Continue w/ antibiotics. Continue pulmonary toileting. Renal failure, acute (Acute renal failure, ARF)
Assessment: Creat 2.0 stable. UOP 35-100ml/hr.
Action: Monitored UOP/ volume status. Recd 2 doses mucomyst pre-cath. Recd IV NaHCO3 bolus and then drip.
Response: Stable Creat 1.8 @ 14:00.
Plan: Continue to monitor.]

**Question:** Is the Nausea condition of this patient which is detected by the 1st occurrence of the word 'nausea' current to the overall note?
**Answer:** No, the nausea condition is not current to the overall note. The patient's nausea is mentioned in the initial presentation, but there is no further mention or discussion of this symptom in the rest of the note. This suggests that the symptom may have resolved or is not a current concern at the time the note was written.
**Analysis:** In the note provided, "nausea" is mentioned just once, specifically in the initial sentence. Its absence in subsequent sections might lead to an interpretation that it was a symptom from a previous encounter, especially when combined with the term "transferred from", which can imply a change in the patient's condition or setting. These factors could have influenced the interpretation by GPT-4, causing it to mistakenly assume that the nausea was not a symptom of the current encounter.

Table S1. Statistical analysis results on the detection labeling task: (a) Two-way ANOVA test results (b) Simple Main Effects Analysis results.

| *Source* | SS | df | MS | F | Prob>F |
|---|---|---|---|---|---|
| **LLMs** | 0.578 | 2 | 0.289 | 21105.54 | $p \ll 0.0001$ |
| **Datasets** | 0.359 | 1 | 0.359 | 26221.67 | $p \ll 0.0001$ |
| **Interaction** | 0.094 | 2 | 0.047 | 3427.72 | $p \ll 0.0001$ |
| **Error** | 0.004 | 294 | 0.000 | | |
| **Total** | 1.035 | 299 | | | |

(a)

| *Group* | Control Group | Lower Limit | Difference | Upper Limit | P-value |
|---|---|---|---|---|---|
| Davinci | GPT-3.5 | -0.055 | -0.054 | -0.052 | $p \ll 0.0001$ |
| Davinci | GPT-4 | -0.059 | -0.057 | -0.056 | $p \ll 0.0001$ |
| GPT-3.5 | GPT-4 | -0.005 | -0.004 | -0.003 | $p \ll 0.0001$ |

(b)

Table S2. Statistical analysis results on the encounter labeling task: (a) Three-way ANOVA test results; (b) Simple Main Effects Analysis results on the concurred dataset; (c) Simple Main Effects Analysis results on the dissenting dataset.

| *Source* | SS | df | MS | F | Prob>F |
|---|---|---|---|---|---|
| **Datasets** | 12.757 | 1 | 12.757 | 52400.96 | 0 |
| **LLMs** | 1.884 | 2 | 0.942 | 3869.33 | 0 |
| **Prompts** | 3.743 | 2 | 1.872 | 7687.62 | 0 |
| **Datasets * LLMs** | 0.039 | 2 | 0.020 | 80.93 | 0 |

| | | | | | | |
|---|---|---|---|---|---|---|
| Datasets * Prompts | 0.058 | 2 | 0.029 | 119.01 | 0 | |
| LLMs * Prompts | 4.012 | 4 | 1.003 | 4119.88 | 0 | |
| Error | 0.216 | 886 | 0.000 | | 0 | |
| Total | 22.709 | 899 | | | | |

(a)

| *Group* | Control Group | Prompt | Lower Limit | Difference | Upper Limit | P-value |
|---|---|---|---|---|---|---|
| Davinci | GPT-3.5 | 1 | 0.270 | 0.273 | 0.276 | p << 0.0001 |
| Davinci | GPT-4 | 1 | -0.102 | -0.099 | -0.096 | p << 0.0001 |
| GPT-3.5 | GPT-4 | 1 | -0.376 | -0.373 | -0.370 | p << 0.0001 |
| Davinci | GPT-3.5 | 2 | 0.004 | 0.006 | 0.007 | p << 0.0001 |
| Davinci | GPT-4 | 2 | 0.055 | 0.057 | 0.058 | p << 0.0001 |
| GPT-3.5 | GPT-4 | 2 | 0.049 | 0.051 | 0.053 | p << 0.0001 |
| Davinci | GPT-3.5 | 3 | -0.024 | -0.023 | -0.021 | p << 0.0001 |
| Davinci | GPT-4 | 3 | -0.024 | -0.023 | -0.022 | p << 0.0001 |
| GPT-3.5 | GPT-4 | 3 | -0.002 | -0.000 | 0.001 | 1 |

(b)

| *Group* | Control Group | Prompt | Lower Limit | Difference | Upper Limit | P-value |
|---|---|---|---|---|---|---|
| Davinci | GPT-3.5 | 1 | 0.216 | 0.221 | 0.227 | p << 0.0001 |
| Davinci | GPT-4 | 1 | -0.055 | -0.049 | -0.044 | p << 0.0001 |
| GPT-3.5 | GPT-4 | 1 | -0.276 | -0.271 | -0.265 | p << 0.0001 |
| Davinci | GPT-3.5 | 2 | 0.045 | 0.050 | 0.054 | p << 0.0001 |
| Davinci | GPT-4 | 2 | 0.069 | 0.074 | 0.079 | p << 0.0001 |
| GPT-3.5 | GPT-4 | 2 | 0.020 | 0.025 | 0.029 | p << 0.0001 |
| Davinci | GPT-3.5 | 3 | 0.032 | 0.037 | 0.042 | p << 0.0001 |
| Davinci | GPT-4 | 3 | 0.002 | 0.007 | 0.011 | 0.0039641 |
| GPT-3.5 | GPT-4 | 3 | -0.036 | -0.031 | -0.026 | p << 0.0001 |

(c)

Table S3. Statistical analysis results on the negation labeling task: (a) Three-way ANOVA test results; (b) Simple Main Effects Analysis results on the concurred dataset; (c) Simple Main Effects Analysis results on the dissenting dataset.

| *Source* | SS | df | MS | F | Prob>F |
|---|---|---|---|---|---|
| Datasets | 5.482 | 1 | 5.482 | 3791.18 | 0 |
| LLMs | 32.242 | 2 | 16.121 | 11148.79 | 0 |
| Prompts | 5.179 | 2 | 2.590 | 1790.91 | 0 |
| Datasets * LLMs | 0.629 | 2 | 0.314 | 217.43 | 0 |
| Datasets * Prompts | 0.682 | 2 | 0.341 | 235.87 | 0 |
| LLMs * Prompts | 10.821 | 4 | 2.705 | 1870.82 | 0 |
| Error | 1.281 | 886 | 0.001 | | 0 |
| Total | 56.316 | 899 | | | |

(a)

| *Group* | Control Group | Prompt | Lower Limit | Difference | Upper Limit | P-value |
|---|---|---|---|---|---|---|
| Davinci | GPT-3.5 | 1 | 0.003 | 0.006 | 0.008 | p << 0.0001 |
| Davinci | GPT-4 | 1 | -0.136 | -0.134 | -0.131 | p << 0.0001 |
| GPT-3.5 | GPT-4 | 1 | -0.142 | -0.139 | -0.137 | p << 0.0001 |

| Group | Control Group | Prompt | Lower Limit | Difference | Upper Limit | P-value |
|---|---|---|---|---|---|---|
| Davinci | GPT-3.5 | 2 | 0.768 | 0.771 | 0.773 | p << 0.0001 |
| Davinci | GPT-4 | 2 | 0.020 | 0.023 | 0.025 | p << 0.0001 |
| GPT-3.5 | GPT-4 | 2 | -0.751 | -0.748 | -0.746 | p << 0.0001 |
| Davinci | GPT-3.5 | 3 | 0.558 | 0.561 | 0.563 | p << 0.0001 |
| Davinci | GPT-4 | 3 | 0.047 | 0.050 | 0.053 | p << 0.0001 |
| GPT-3.5 | GPT-4 | 3 | -0.514 | -0.511 | -0.508 | p << 0.0001 |

(b)

| *Group* | Control Group | Prompt | Lower Limit | Difference | Upper Limit | P-value |
|---|---|---|---|---|---|---|
| Davinci | GPT-3.5 | 1 | 0.097 | 0.103 | 0.108 | p << 0.0001 |
| Davinci | GPT-4 | 1 | -0.125 | -0.120 | -0.115 | p << 0.0001 |
| GPT-3.5 | GPT-4 | 1 | -0.228 | -0.223 | -0.218 | p << 0.0001 |
| Davinci | GPT-3.5 | 2 | 0.600 | 0.606 | 0.611 | p << 0.0001 |
| Davinci | GPT-4 | 2 | 0.062 | 0.067 | 0.072 | p << 0.0001 |
| GPT-3.5 | GPT-4 | 2 | -0.544 | -0.538 | -0.533 | p << 0.0001 |
| Davinci | GPT-3.5 | 3 | 0.261 | 0.266 | 0.271 | p << 0.0001 |
| Davinci | GPT-4 | 3 | -0.075 | -0.071 | -0.066 | p << 0.0001 |
| GPT-3.5 | GPT-4 | 3 | -0.341 | -0.336 | -0.332 | p << 0.0001 |

(c)

Table S4. Simple Main Effects Analysis results between zero-shot and fine-tuned models: (a) on the detection labeling task; (b) on the encounter labeling task; (c) on the negation labeling task.

| *Group* | Control Group | Lower Limit | Difference | Upper Limit | P-value |
|---|---|---|---|---|---|
| Davinci (zero-shot) | GPT-3.5 (zero-shot) | -0.068 | -0.067 | -0.065 | p << 0.0001 |
| Davinci (zero-shot) | GPT-4 (zero-shot) | -0.075 | -0.074 | -0.072 | p << 0.0001 |
| Davinci (zero-shot) | LLaMA (zero-shot) | -0.056 | -0.054 | -0.053 | p << 0.0001 |
| Davinci (zero-shot) | LLaMA (fine-tuned) | -0.001 | 0.001 | 0.002 | 1 |
| GPT-3.5 (zero-shot) | GPT-4 (zero-shot) | -0.009 | -0.007 | -0.005 | p << 0.0001 |
| GPT-3.5 (zero-shot) | LLaMA 2 (zero-shot) | 0.011 | 0.012 | 0.014 | p << 0.0001 |
| GPT-3.5 (zero-shot) | LLaMA 2 (fine-tuned) | 0.066 | 0.067 | 0.069 | p << 0.0001 |
| GPT-4 (zero-shot) | LLaMA 2 (zero-shot) | 0.018 | 0.019 | 0.021 | p << 0.0001 |
| GPT-4 (zero-shot) | LLaMA 2 (fine-tuned) | 0.073 | 0.074 | 0.076 | p << 0.0001 |
| LLaMA 2 (zero-shot) | LLaMA 2 (fine-tuned) | 0.053 | 0.055 | 0.057 | p << 0.0001 |

(a)

| *Group* | Control Group | Lower Limit | Difference | Upper Limit | P-value |
|---|---|---|---|---|---|
| Davinci (zero-shot) | GPT-3.5 (zero-shot) | 0.256 | 0.260 | 0.263 | p << 0.0001 |
| Davinci (zero-shot) | GPT-4 (zero-shot) | -0.070 | -0.066 | -0.063 | p << 0.0001 |
| Davinci (zero-shot) | LLaMA (zero-shot) | -0.062 | -0.058 | -0.055 | p << 0.0001 |
| Davinci (zero-shot) | LLaMA (fine-tuned) | -0.073 | -0.069 | -0.066 | p << 0.0001 |
| GPT-3.5 (zero-shot) | GPT-4 (zero-shot) | -0.330 | -0.326 | -0.323 | p << 0.0001 |
| GPT-3.5 (zero-shot) | LLaMA 2 (zero-shot) | -0.322 | -0.318 | -0.315 | p << 0.0001 |
| GPT-3.5 (zero-shot) | LLaMA 2 (fine-tuned) | -0.332 | -0.330 | -0.326 | p << 0.0001 |
| GPT-4 (zero-shot) | LLaMA 2 (zero-shot) | 0.005 | 0.008 | 0.011 | p << 0.0001 |
| GPT-4 (zero-shot) | LLaMA 2 (fine-tuned) | -0.006 | -0.003 | 0.001 | 0.199 |

| Group | Control Group | Lower Limit | Difference | Upper Limit | P-value |
|---|---|---|---|---|---|
| LLaMA 2 (zero-shot) | LLaMA 2 (fine-tuned) | -0.014 | -0.011 | -0.007 | p << 0.0001 |

(b)

| Group | Control Group | Lower Limit | Difference | Upper Limit | P-value |
|---|---|---|---|---|---|
| Davinci (zero-shot) | GPT-3.5 (zero-shot) | 0.047 | 0.051 | 0.055 | p << 0.0001 |
| Davinci (zero-shot) | GPT-4 (zero-shot) | -0.122 | -0.118 | -0.115 | p << 0.0001 |
| Davinci (zero-shot) | LLaMA (zero-shot) | -0.027 | -0.023 | -0.020 | p << 0.0001 |
| Davinci (zero-shot) | LLaMA (fine-tuned) | 0.120 | 0.123 | 0.127 | p << 0.0001 |
| GPT-3.5 (zero-shot) | GPT-4 (zero-shot) | -0.173 | -0.169 | -0.166 | p << 0.0001 |
| GPT-3.5 (zero-shot) | LLaMA 2 (zero-shot) | -0.078 | -0.074 | -0.071 | p << 0.0001 |
| GPT-3.5 (zero-shot) | LLaMA 2 (fine-tuned) | 0.069 | 0.072 | 0.076 | p << 0.0001 |
| GPT-4 (zero-shot) | LLaMA 2 (zero-shot) | 0.091 | 0.095 | 0.099 | p << 0.0001 |
| GPT-4 (zero-shot) | LLaMA 2 (fine-tuned) | 0.238 | 0.242 | 0.245 | p << 0.0001 |
| LLaMA 2 (zero-shot) | LLaMA 2 (fine-tuned) | 0.143 | 0.147 | 0.150 | p << 0.0001 |

(c)